\documentclass{article} 
\usepackage{iclr2027_conference,times}

\usepackage{amsmath,amsfonts,bm}

\def\eqref#1{equation~\ref{#1}}

\def\1{\bm{1}}

\DeclareMathAlphabet{\mathsfit}{\encodingdefault}{\sfdefault}{m}{sl}
\SetMathAlphabet{\mathsfit}{bold}{\encodingdefault}{\sfdefault}{bx}{n}

\usepackage{hyperref}
\usepackage{url}
\usepackage{booktabs}
\usepackage{graphicx} 
\usepackage{amssymb} 
\usepackage{wrapfig}

\title{SuperMotion: Source-Preserving Denoising for Text-Driven Human Motion Editing}

\author{Fa-Ting Hong, Peter Wonka\\
King Abdullah University of Science and Technology \\
\texttt{\{fating.hong, peter.wonka\}@kaust.edu.sa} \\
}

\iclrfinalcopy 
\begin{document}

\maketitle
\begin{abstract}

Text-driven human motion editing aims to realize a requested change while preserving compatible source content. Existing diffusion editors rely largely on learned conditioning for preservation of the unedited part, yet their outputs can lose temporal detail as denoising proceeds. We propose the \textbf{Source-Preserving Denoising framework (SuperMotion)}, which explicitly reuses the source at each reverse step for source preservation. We first align the source motion to the output timeline and predict a preservation gate that controls reuse across frames and feature dimensions. A clean-space source anchor then utilizes the learned preservation gate to blend the predicted clean motion with the aligned source and passes the corrected estimate directly to the sampling posterior. Because the aligned source is a realized motion rather than a regression output, the anchor injects sample-level temporal detail that a reconstruction-trained denoiser tends to smooth away. To learn effective source reuse, we supervise the anchored estimate against the editing target and match its second temporal differences through a temporal high-frequency loss. These objectives require no explicit edit masks. Extensive experiments show that SuperMotion improves editing accuracy, reaching 33.20\% full-pool R@1 on MotionFix, while reducing temporal-detail attenuation and preserving motion dynamics as it realizes the requested changes. Ablations confirm that the learned preservation gate is responsible for the gain and that it reuses the source to retain the unedited content properly. Project page: \url{https://harlanhong.github.io/Projects/SuperMotion/index.html}

\end{abstract}
\section{Introduction}\label{sec:introduction}
\begin{figure}[!ht]
    \centering
    \includegraphics[width=\linewidth]{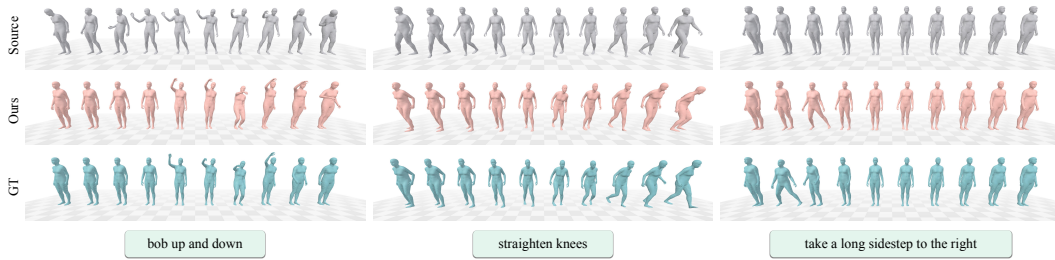}
    \vspace{-20pt}
    \caption{\textbf{Motion editing requires both change and preservation.}
    Given a source motion (gray), the edit should realize the requested modification while retaining compatible source content. Our outputs (salmon) and reference targets (turquoise) illustrate three instructions, ``bob up and down,'' ``straighten knees,'' and ``take a long sidestep to the right.'' Each strip shows successive poses.}
    \label{fig:editing-teaser}
\end{figure}

Text-driven human motion editing modifies an existing movement according to a textual instruction. A request such as ``straighten knees'' specifies a change while preserving other aspects of the source movement (Figure~\ref{fig:editing-teaser}). A successful edit must therefore balance instruction compliance with preservation of compatible source content. This balance requires more than copying a predefined region. Source and reference target motions may differ in duration, coordination, and execution, so content that should be retained cannot be preserved exactly, but only approximately. The editor must determine how to use the source while accommodating the requested change.

Advances in text-conditioned motion generation provide a foundation for motion editing~\citep{petrovich2022temos,tevet2022human,zhang2024motiondiffuse,guo2024momask}. Existing editors build on these models by improving how they use the source motion and instruction, learning motion similarity, and introducing objectives for editing body parts and preserving source content~\citep{athanasiou2024motionfix,li2025simmotionedit,guo2025motionlab,yang2025partmotionedit,shi2026omni}. These methods can produce the requested action, but the resulting movement may be too subdued or too exaggerated compared with the reference target. Retrieval scores measure how well the generated motion matches the target in a learned representation, but do not directly measure these differences in movement. We therefore examine how temporal detail changes in the generated motion and throughout denoising.

\begin{wrapfigure}{r}{0.6\textwidth}
    \centering
    \vspace{-10pt}
    \includegraphics[width=\linewidth]{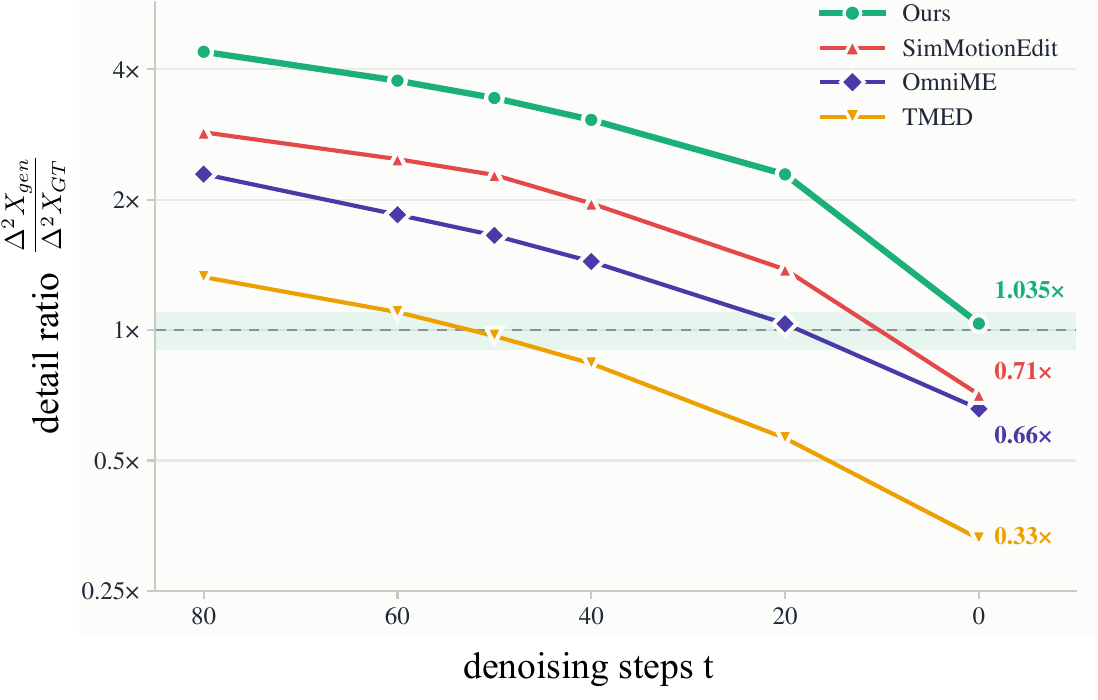}
    \vspace{-20pt}
    \caption{\textbf{Detail attenuation during denoising.}
    Median generated-to-target second-order difference magnitude at selected timesteps, with denoising proceeding from left to right. The plotted baselines finish below the target reference of one, while the shown SuperMotion run ends at $1.035\times$. Shading marks a fixed reference band from $0.9$ to $1.1$. Agreement in this statistic alone does not establish a correct edit.}
    \label{fig:detail-denoising}
    \vspace{-8pt}
\end{wrapfigure}

Concretely, we define a \emph{detail ratio}: the magnitude of second-order temporal differences of the running clean-motion estimate \(\hat{x}_0\) at a given reverse step, normalized by the same statistic computed on the ground-truth target. A well-calibrated reverse process should drive this ratio toward \(1\times\) and hold it there. Figure~\ref{fig:detail-denoising} shows that prior diffusion editors behave otherwise: the median ratio decreases along the reverse trajectory, crosses \(1\times\), and keeps falling, reaching \(0.33\times\) for TMED~\citep{athanasiou2024motionfix}, \(0.66\times\) for OmniME~\citep{shi2026omni}, and \(0.71\times\) for SimMotionEdit~\citep{li2025simmotionedit}. The decline concentrates in the final low-noise steps, where the estimate already matches the target detail level and further denoising suppresses local temporal variation instead of refining it.
A denoiser trained with a reconstruction objective approximates a posterior mean~\citep{efron2011tweedie}, and such distortion-optimal estimates are necessarily smoother than individual samples~\citep{blau2018perception}. The effect is especially costly in motion editing. The content that should be preserved already exists in the source, but regenerating the sequence from noise forces the model to resynthesize it, after which it is pulled toward the mean. 

These observations motivate explicit control over source reuse during denoising. In frameworks that reconstruct the target from noise, source conditioning provides information about the input motion, but retaining its content still depends on the denoiser's predictions~\citep{athanasiou2024motionfix,li2025simmotionedit}. The source could also provide a clean signal against which those predictions are adjusted at each reverse step. Related ideas appear in diffusion-based image editing and inpainting, where spatial masks or known regions guide the preservation of existing content~\citep{avrahami2022blended,lugmayr2022repaint,couairon2022diffedit,zhang2025semantic}. Applying this principle to motion editing requires accounting for differences between the source and target sequences. Because no preservation mask is supplied, the model must learn how strongly to reuse the aligned source while allowing the instruction to change the movement.

We introduce the \textbf{S}o\textbf{u}rce-\textbf{P}res\textbf{er}ving Denoising framework (SuperMotion) to learn this balance within the reverse process. Its central component, a \emph{Clean-Space Source Anchor}, blends the current clean-motion estimate with a temporally aligned source through a learned \emph{Source Preservation Gate}. The gate assigns a reuse weight to each frame and feature dimension at each denoising step. Applied after classifier-free guidance and clean-motion reconstruction, the anchor produces the estimate used directly by the sampling posterior. 
This placement gives the gate a consistent interpretation as a convex blending weight and avoids the inverse noise-scale factor required to express the same clean-motion correction in velocity space. 
Supervision against the editing target trains the gate to adjust source reuse without an explicit edit mask. We train SuperMotion with a source-anchor loss that learns source reuse from the editing target and a temporal high-frequency loss that matches second temporal differences. 

We conduct extensive experiments to evaluate our method. The results show that SuperMotion improves instruction compliance while reducing temporal-detail attenuation. Ablation studies show that supervising the anchored prediction against the editing target is critical to learning effective source reuse. Our contributions are threefold.
\begin{enumerate}
    \item We identify temporal-detail attenuation as a systematic failure of diffusion motion editors. Their clean-motion estimates lose second-order temporal variation during late denoising, and several published editors score below a copy-source baseline.
    \item We propose SuperMotion, which reuses a temporally aligned source at each reverse step through a learned per-frame, per-dimension gate on the clean-motion estimate, trained from the editing target without edit masks.
   \item SuperMotion reaches 79.17\% batch R@1 and 33.20\% full-pool R@1 on MotionFix and 46.88\% batch R@1 on STANCE Adjustment, surpassing the strongest prior editor by 1.88, 1.18, and 3.13 points. It restores the detail ratio to 1.035, against 0.66 to 0.71 for prior editors, confirming that the learned preservation gate recovers high-frequency detail from the source motion.
\end{enumerate}
\section{Related Work}\label{sec:related}

\noindent\textbf{Human Motion Generation.}
Text-driven human motion generation has progressed through
continuous latent modeling, diffusion-based synthesis, and
discrete motion modeling.
Guo et al.~\citep{guo2022generating} investigate diverse
text-conditioned motion synthesis and introduce HumanML3D,
supporting subsequent advances in language-driven animation.
TEMOS~\citep{petrovich2022temos} learns compatible text and motion
distributions through variational training, while
MotionCLIP~\citep{tevet2022motionclip} aligns motion representations
with the CLIP embedding space.
Diffusion-based methods, including MDM~\citep{tevet2022human},
MotionDiffuse~\citep{zhang2024motiondiffuse}, and FLAME~\citep{kim2023flame},
generate motions through text-conditioned iterative denoising.
MLD~\citep{chen2023executing} instead performs diffusion in a compact
latent space to reduce computational cost. Discrete motion representations provide an alternative to
continuous-space generation.
T2M-GPT~\citep{zhang2023t2mgpt} combines vector quantization with
autoregressive token prediction, while
MotionGPT~\citep{jiang2023motiongpt} models motion and language within
a unified token-based framework.
MoMask~\citep{guo2024momask} and MMM~\citep{pinyoanuntapong2024mmm} adopt masked motion
modeling, enabling iterative token reconstruction and flexible
generation.
Beyond global text conditioning, FineMoGen~\citep{zhang2023finemogen}
introduces fine-grained spatio-temporal control, and
MaskControl~\citep{pinyoanuntapong2025maskcontrol} incorporates spatial constraints
into masked motion generation.
Together, these approaches improve the diversity, efficiency,
and controllability of human motion synthesis.

Our focus is source-conditioned editing rather than synthesizing
a new realization from text alone.
SuperMotion retains a diffusion backbone and introduces a learned
source-reuse operator to preserve compatible motion content
during denoising.

\noindent\textbf{Human Motion Editing.}
Human motion editing modifies an existing sequence while
retaining content compatible with the requested change.
Early generative approaches enable editing through semantic
latent manipulation or constrained reconstruction.
MotionCLIP~\citep{tevet2022motionclip} supports editing in a
language-aligned latent space, while MDM~\citep{tevet2022human} and
FLAME~\citep{kim2023flame} support motion completion and editing
through constrained diffusion.
MMM~\citep{pinyoanuntapong2024mmm} reconstructs masked motion tokens, and
FineMoGen~\citep{zhang2023finemogen} supports fine-grained editing
through spatio-temporal descriptions.
These approaches establish useful editing capabilities,
although their control interfaces and supervision differ
from paired instruction-based editing. MotionFix~\citep{athanasiou2024motionfix} establishes a benchmark of
source--instruction--target triplets and introduces a
source-conditioned diffusion editor.
Subsequent work improves instruction alignment, motion
preservation, and editing versatility.
SimMotionEdit~\citep{li2025simmotionedit} jointly trains motion
editing and motion-similarity prediction, while
PartMotionEdit~\citep{yang2025partmotionedit} combines part-aware
modulation with part-level similarity supervision and
bidirectional text--motion interaction.
MotionReFit~\citep{jiang2025dynamic} expands editing supervision
through MotionCutMix and uses an autoregressive diffusion
model with a motion coordinator.
MotionLab~\citep{guo2025motionlab} unifies generation and editing
through a motion--condition--motion formulation based on
rectified flows.
OmniME~\citep{shi2026omni} combines intermediate-feature,
preservation, and positive--negative semantic supervision.
Recent work further considers both spatial preservation
and temporal rhythm through non-uniform temporal
alignment~\citep{lin2026spatial}.

SuperMotion uses a learned, feature-wise source anchor to modify
the clean-motion estimate after guidance and before each
posterior update.
Its distinction lies in this explicit sampling-time operator,
rather than preservation supervision or source blending
as a general objective.
\section{Method}\label{sec:method}
\begin{figure}[t]
    \centering
    \includegraphics[width=\linewidth]{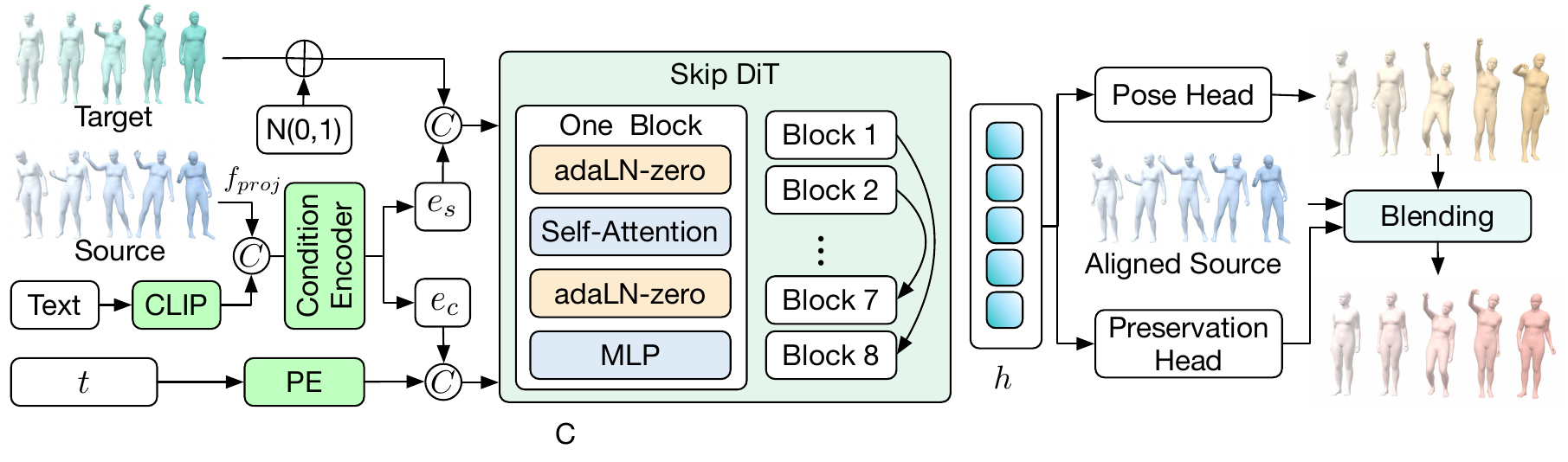}
    \vspace{-20pt}
    \caption{Overall architecture. A Skip DiT takes the noised motion $x_t$, a condition encoder over source and text, and the diffusion timestep. Shared target tokens feed a pose head that predicts velocity $v$ and a preservation head that predicts a per-frame, per-dimension gate $g$. After converting $v$ to $\hat{x}_0$, blending pulls the estimate toward the time-aligned source where $g$ is high and leaves it to the pose head where $g$ is low.}
    \label{fig:framework}
\end{figure}

Given a source motion and an editing instruction, our goal is to generate a motion that realizes the requested change while retaining compatible source content. Our Source-Preserving Denoising (SuperMotion) makes source reuse an explicit operation inside the reverse process, applied to the clean-motion estimate at every step and controlled by a gate learned from the editing target alone.

\subsection{Problem Formulation}
Text-based 3D motion editing takes a source sequence $s\in\mathbb{R}^{T_s\times D}$ and a natural-language instruction $c$, and produces a motion $\hat{x}_0\in\mathbb{R}^{T\times D}$ that carries out $c$. We use the MotionFix feature of $D{=}207$ z-normalized dimensions per frame (pelvis translation increments, 6D orientation, $z$-orientation increments, 21-joint body pose, and 22 local joint coordinates) at 30\,fps.

The instruction typically changes only part of the source. Over the remaining frames and coordinates the target coincides with the source, so the underlying edit is close to an identity mapping. Diffusion editors nevertheless regenerate the full sequence from Gaussian noise with $s$ and $c$ used only as conditions, so every reverse step re-synthesizes coordinates that could have been copied. A denoiser trained with a reconstruction loss approximates the posterior mean $\mathbb{E}[x_0\mid x_t]$, and such distortion-optimal estimates are smoother than any individual sample~\citep{blau2018perception}. The identity part of the edit is therefore re-estimated from noise at every step and pulled toward a mean it does not belong to.

Let $x_0$ be the clean target motion and
\begin{equation}
x_t=\sqrt{\bar\alpha_t}\,x_0+\sqrt{1-\bar\alpha_t}\,\epsilon,\qquad\epsilon\sim\mathcal{N}(0,I)
\end{equation}
the corresponding noised state at diffusion time $t$. A reverse step must recover a clean estimate from $x_t$ that realizes $c$ and remains close to the source wherever $c$ does not ask for a change. SuperMotion enforces the second requirement as an operation on the clean estimate rather than leaving it to conditioning.

\subsection{Architecture Overview}
As shown in Figure~\ref{fig:framework}, our implementation builds on SimMotionEdit~\citep{li2025simmotionedit}. A four-layer Transformer condition encoder maps the linear projection of source motion $s$ and the CLIP ViT-L/14 text embedding~\citep{radford2021learning} of $c$ to token sequences $e_s$ and $e_c$, respectively. The projected noisy-motion tokens are concatenated with $e_s$ and processed by a diffusion Transformer of eight DiT blocks~\citep{peebles2023scalable} with hidden width $512$ and long skip connections between symmetric blocks following U-ViT~\citep{bao2023all}. After removing the source tokens, the target features $h\in\mathbb{R}^{T\times512}$ feed two heads. All quantities computed from $x_t$ depend on the timestep, which we leave implicit.

The pose head outputs $\hat{v}\in\mathbb{R}^{T\times D}$ under the $v$-parameterization~\citep{salimans2022progressive}, giving the clean-motion estimate
\begin{equation}
    \hat{x}_0
    =
    \sqrt{\bar{\alpha}_t}\,x_t
    -
    \sqrt{1-\bar{\alpha}_t}\,\hat{v}.
    \label{eq:v-to-clean}
\end{equation}
Unlike SimMotionEdit, which predicts $x_0$ directly, we predict velocity $v$. The velocity loss equals $(\mathrm{SNR}(t)+1)\,\|x_0-\hat{x}_0\|_2^2$ with $\mathrm{SNR}(t)=\bar{\alpha}_t/(1-\bar{\alpha}_t)$~\citep{kingma2021vdm}, and therefore places more weight on clean-motion errors at low noise, where the detail loss of Section~\ref{sec:introduction} concentrates. The preservation head, an MLP on the same $h$, outputs the preservation gate, which indicates how much of each frame and pose coordinate should be retrieved from the aligned source. Then, we use the predicted gate value to blend the estimation $\hat{x}_0$ with the aligned source motion (Section~\ref{sec:spd}).

At inference, we use classifier-free guidance~\citep{ho2022classifier} with separate source and text scales, following MotionFix~\citep{athanasiou2024motionfix}.
\begin{equation}
    \hat{v}_{\mathrm{cfg}}
    =
    \hat{v}_{\varnothing}
    +
    w_s\bigl(\hat{v}_{s}-\hat{v}_{\varnothing}\bigr)
    +
    w_c\bigl(\hat{v}_{s,c}-\hat{v}_{s}\bigr),
    \label{eq:cfg}
\end{equation}
where $w_s$ and $w_c$ scale the source and text guidance, respectively. We convert $\hat{v}_{\mathrm{cfg}}$ to $\hat{x}_0$ by Eq.~\ref{eq:v-to-clean} and apply the anchor with the gate from the fully conditioned branch.

\subsection{Source-Preserving Denoising}
\label{sec:spd}

\paragraph{Aligning the source.}
Source and target may differ in duration. We linearly resample the source onto the output timeline,
\begin{equation}
    \bar{s}=\operatorname{Align}(s,T)\in\mathbb{R}^{T\times D},
\end{equation}
which yields a reference of the same length as $\hat{x}_0$ without assuming that corresponding frames contain identical movements. Residual misalignment is absorbed by the gate.
\paragraph{Preservation gate.}
From the target-token features, the preservation head predicts
\begin{equation}
    g=\operatorname{sigmoid}\bigl(f_\phi(h)\bigr),
    \qquad
    g\in(0,1)^{T\times D},
    \label{eq:preservation-field}
\end{equation}
where $f_\phi$ is an MLP. Each entry is the source-reuse weight for one frame and one feature dimension, and the dependence on $t$ through $h$ lets reuse vary along the reverse trajectory.

\paragraph{Clean-space source anchor.}
The anchor forms the corrected estimate
\begin{equation}
    \widetilde{x}_0
    =
    \hat{x}_0
    +
    g\odot(\bar{s}-\hat{x}_0)
    =
    (1-g)\odot\hat{x}_0+g\odot\bar{s}.
    \label{eq:clean-source-anchor}
\end{equation}
Because $g\in(0,1)$, every corrected coordinate lies between the denoiser prediction and the aligned source. The anchor addresses the detail loss of Section~\ref{sec:introduction} directly. The learned estimate $\hat{x}_0$ is trained as a regressor and inherits the smoothing of a posterior mean, whereas $\bar{s}$ is a realized motion and carries sample-level temporal detail. On coordinates the instruction leaves unchanged, $\bar{s}$ is close to $x_0$, so Eq.~\ref{eq:clean-source-anchor} injects detail the target should have without asking the network to regenerate it. Conditioning provides the same information, but only as network input. Recovering the source then requires the denoiser to learn an identity map through its full depth, which Figure~\ref{fig:detail-denoising} in Section~\ref{sec:introduction} shows it does not reliably do. The anchor bypasses this by blending the estimate with the source at a learned, per-coordinate strength.
\paragraph{Optimal gate.}
Holding $\hat{x}_0$ fixed and minimizing the anchor loss of Eq.~\ref{eq:source-anchor-loss} over $g\in(0,1)^{T\times D}$, the closure of the sigmoid range, decouples across coordinates. Each coordinate is a convex quadratic in $g_{i,d}$, so its constrained minimizer is the projection of the unconstrained one onto $[0,1]$,
\begin{equation}
    g^{\ast}_{i,d}
    =
    \operatorname{clip}\!\left(
    \frac{x_{0,i,d}-\hat{x}_{0,i,d}}{\bar{s}_{i,d}-\hat{x}_{0,i,d}},
    \,0,\,1\right),
    \label{eq:gstar}
\end{equation}
defined wherever $\bar{s}_{i,d}\neq\hat{x}_{0,i,d}$ (elsewhere the anchor is inactive and $g_{i,d}$ is immaterial). The sigmoid gate takes values in (0, 1), so it approaches an optimum at $0$ or $1$ without reaching it. The unclipped ratio is the fraction of the way from the denoiser prediction to the source at which the target lies. It equals one where the target coincides with the source and zero where it coincides with the prediction. The bound $(0,1)$ is a design choice and we ablate other choices in Sec.~\ref{sec:ablation}. Where the target lies outside the segment between prediction and source, the gate saturates and the residual is left to the denoiser. Within this bound, Eq.~\ref{eq:gstar} makes two predictions about the learned gate. First, writing the edit as $\delta=x_0-\bar{s}$ gives $g^{\ast}=\operatorname{clip}\bigl(1+\delta/(\bar{s}-\hat{x}_0),\,0,\,1\bigr)$, so an edit can only lower the optimum below one and never raise it, and larger edits tend to yield a smaller optimal gate. Second, the optimum vanishes as $t\to0$. The denoiser prediction approaches the target, the numerator goes to zero wherever $\bar{s}\neq x_0$, and the gate releases the source in the final steps regardless of how much it reused earlier. The learned gate reproduces both behaviors.

\paragraph{Anchoring within reverse diffusion.}
The corrected estimate replaces the clean target in the Gaussian posterior,
\begin{equation}
    p_{\theta,\phi}(x_{t-1}\mid x_t,s,c)
    =
    q(x_{t-1}\mid x_t,\widetilde{x}_0),
    \label{eq:anchored-transition}
\end{equation}
so source content enters the sample at every reverse step rather than only through the network input. We apply the correction to $\hat{x}_0$ rather than to $\hat{v}$. The two are related by a factor $(1-\bar{\alpha}_t)^{-1/2}$ that grows without bound as $t\to0$. Anchoring in clean space keeps $g$ a bounded convex weight at every timestep.

\subsection{Training Objective}
\label{sec:training-objective}

In addition to the velocity prediction loss, we apply two losses to the anchored estimate $\widetilde{x}_0$. We write them for a single target $x_0\in\mathbb{R}^{T\times D}$ with $T$ valid frames, index frames by $i$ and dimensions by $d$, and exclude padded frames.

\paragraph{Source-anchor loss.}
We supervise the corrected prediction against the target,
\begin{equation}
    \mathcal{L}_{\mathrm{anc}}
    =
    \frac{1}{TD}
    \sum_{i=1}^{T}
    \left\|
    \widetilde{x}_{0,i}-x_{0,i}
    \right\|_2^2.
    \label{eq:source-anchor-loss}
\end{equation}
Because $\widetilde{x}_0$ depends on both heads, this loss trains the denoiser and the gate jointly, and its per-coordinate minimizer over $g$ is Eq.~\ref{eq:gstar}. The editing target thus supervises source reuse without an explicit preservation mask.

\paragraph{Temporal high-frequency loss.}
Reconstruction alone does not constrain local temporal variation. We match second-order temporal differences between the anchored prediction and the target. With $\Delta^2 x_i=x_{i+1}-2x_i+x_{i-1}$,
\begin{equation}
    \mathcal{L}_{\mathrm{hf}}
    =
    \frac{1}{(T-2)D}
    \sum_{i=2}^{T-1}
    \left\|
    \Delta^2\widetilde{x}_{0,i}
    -
    \Delta^2x_{0,i}
    \right\|_2^2.
    \label{eq:temporal-hf-loss}
\end{equation}
This term penalizes both excess and deficit of variation at each frame and dimension, and its gradient reaches the gate through the anchor.

\paragraph{Overall objective.}
We retain the velocity loss $\mathcal{L}_v$, a masked mean squared error within each feature group averaged across groups, and the auxiliary similarity loss $\mathcal{L}_{\mathrm{cls}}$ from SimMotionEdit~\citep{li2025simmotionedit}. The complete objective is
\begin{equation}
    \mathcal{L}
    =
    \mathcal{L}_v
    +
    \lambda_{\mathrm{anc}}\mathcal{L}_{\mathrm{anc}}
    +
    \lambda_{\mathrm{hf}}\mathcal{L}_{\mathrm{hf}}
    +
    \lambda_{\mathrm{cls}}\mathcal{L}_{\mathrm{cls}},
    \label{eq:spd-training-objective}
\end{equation}
with $\lambda_{\mathrm{anc}}=3$, $\lambda_{\mathrm{hf}}=0.5$, and $\lambda_{\mathrm{cls}}=0.05$. Losses are aggregated over valid frames and averaged over sampled timesteps and noise.
\section{Experiment}\label{sec:exp}
\begin{table}[t]
\centering
\vspace{-12pt}
\caption{Comparison on MotionFix.}
\small
\setlength{\tabcolsep}{4pt}
\resizebox{\textwidth}{!}{
\begin{tabular}{l|cccccccc}
\toprule
Method & B R@1$\uparrow$ & B R@2$\uparrow$ & B R@3$\uparrow$ & B AvgR$\downarrow$ & FR@1$\uparrow$ & FR@2$\uparrow$ & FR@3$\uparrow$ & F AvgR$\downarrow$ \\
\midrule
Copy source & 74.79 & 85.42 & 91.25 & 1.88 & 30.24 & 44.47 & 54.35 & 15.53 \\
MDM~\citep{tevet2022human} & 4.03 & 7.56 & 10.48 & 15.55 & 0.10 & 0.10 & 0.10 & - \\
MDM-BP~\citep{tevet2022human}  & 39.10 & 50.09 & 54.84 & 6.46 & 8.69 & 14.71 & 18.36 & 180.99 \\
MotionLab~\citep{guo2025motionlab} & 54.53 & 70.02 & 77.38 & 3.42 & - & - & - & - \\
TMED~\citep{athanasiou2024motionfix} & 62.90 & 76.51 & 83.06 & 2.71 & 14.51 & 21.72 & 28.73 & 56.63 \\
MotionReFit~\citep{jiang2025dynamic} & 66.63 & 80.05 & 84.98 & 2.64 & 14.13 & 23.52 & 30.53 & 54.06 \\
SimMotionEdit~\citep{li2025simmotionedit} & 70.62 & 82.92 & 88.12 & 2.38 & 25.49 & 39.33 & 49.21 & 23.49 \\
PartMotionEdit~\citep{yang2025partmotionedit} & 73.96 & 85.83 & 90.21 & 1.92 & 27.27 & 45.06 & 53.36 & 16.24 \\
Han et al.~\citep{han2026cross} & 74.38 & 88.54 & 92.08 & 1.92 & 29.45 & 45.26 & 54.55 & 16.42 \\
OmniME~\citep{shi2026omni} & 77.29 & 88.54 & 91.88 & 1.79 & 32.02 & 50.20 & \textbf{59.88} & 13.06 \\
\midrule
\textbf{Ours} & \textbf{79.17} & \textbf{89.17} & \textbf{92.92} & \textbf{1.68} & \textbf{33.20} & \textbf{50.79} & 59.09 & \textbf{12.87} \\

\bottomrule
\end{tabular}
}
\label{tab:motionfix}
\end{table}

\begin{table}[t]
\centering
\vspace{-12pt}
\caption{Comparison on STANCE Adjustment.}
\small
\setlength{\tabcolsep}{4pt}
\resizebox{\textwidth}{!}{
\begin{tabular}{l|cccccccc}
\toprule
Method & B R@1$\uparrow$ & B R@2$\uparrow$ & B R@3$\uparrow$ & B AvgR$\downarrow$ & FR@1$\uparrow$ & FR@2$\uparrow$ & FR@3$\uparrow$ & F AvgR$\downarrow$ \\
\midrule
Copy source & 40.10 & 58.33 & 64.58 & 5.13 & 17.35 & 22.96 & 30.10 & 27.23 \\
MotionReFit~\citep{jiang2025dynamic} & 25.52 & 34.90 & 45.83 & 8.27 & 9.18 & 14.29 & 16.84 & 48.74 \\
TMED~\citep{athanasiou2024motionfix} & 39.58 & 49.48 & 60.94 & 6.26 & 13.78 & 22.96 & 30.10 & 34.38 \\
SimMotionEdit~\citep{li2025simmotionedit} & 43.23 & 59.38 & 68.23 & 4.67 & 20.92 & \textbf{31.63} & 37.24 & 24.02 \\
OmniME~\citep{shi2026omni} & 43.75 & 56.25 & 66.15 & 4.66 & \textbf{22.45} & \textbf{31.63} & 36.22 & 22.77 \\
\midrule
\textbf{Ours} & \textbf{46.88} & \textbf{63.02} & \textbf{71.88} & \textbf{3.72} & 18.37 & 31.12 & \textbf{38.27} & \textbf{19.21} \\

\hline
\end{tabular}
}
\label{tab:stance}
    \vspace{-15pt}
\end{table}
\begin{figure}[ht]
    \centering
    \includegraphics[width=\linewidth]{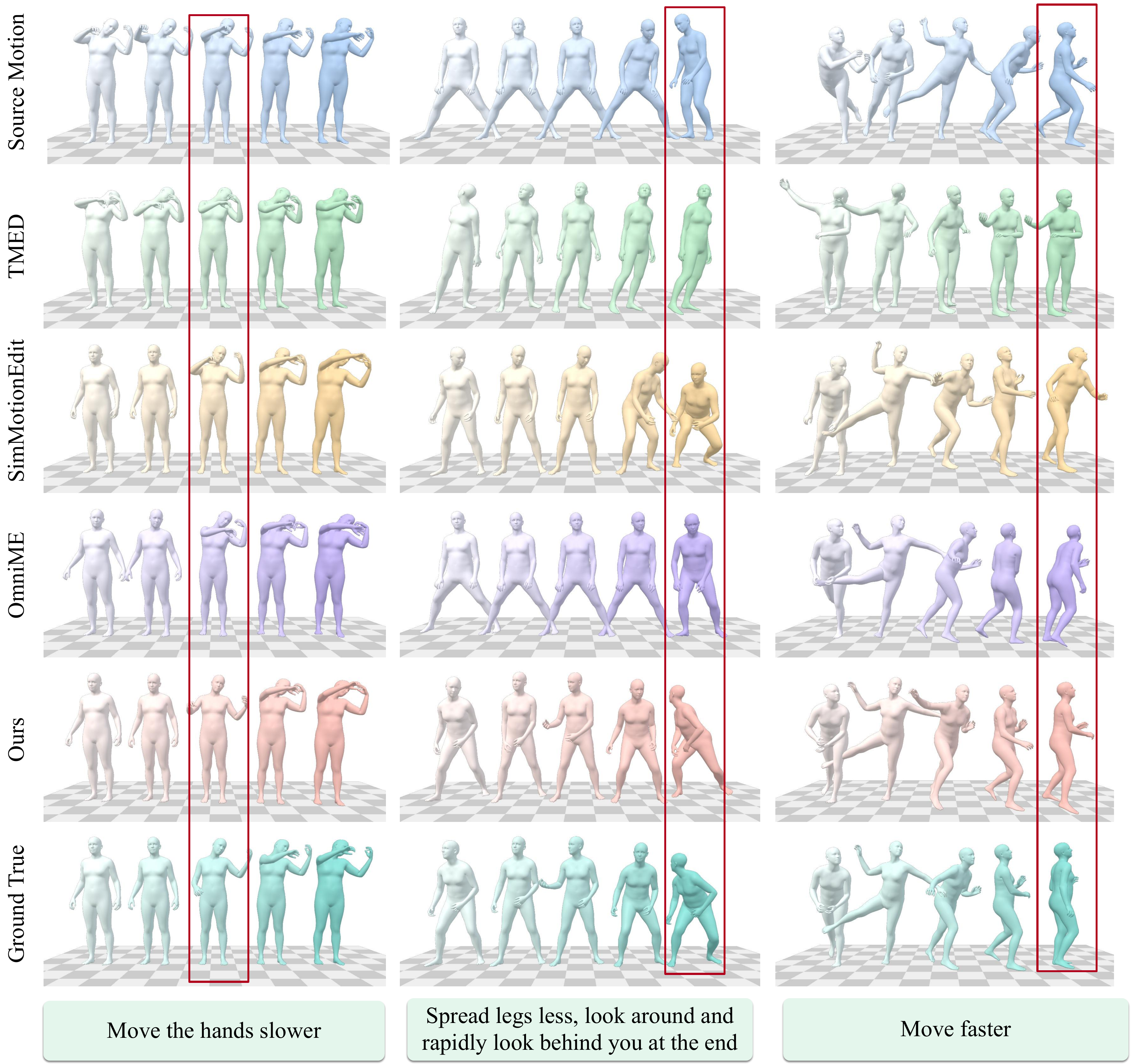}
    \vspace{-15pt}
    \caption{\textbf{Qualitative comparison on MotionFix dataset.} Relative to prior methods, ours follows the instruction on the first two examples and keeps the unedited motion on the third.}
    \vspace{-15pt}
    
\label{fig:qual_motionfix}
\end{figure}

\begin{table}[t]
\centering
\vspace{-12pt}
\caption{Ablation study on different components (MotionFix).}
\small
\setlength{\tabcolsep}{4pt}
\begin{tabular}{l|cccccccc}
\hline
Method & B R@1$\uparrow$ & B R@2$\uparrow$ & B R@3$\uparrow$ & B AvgR$\downarrow$ & FR@1$\uparrow$ & FR@2$\uparrow$ & FR@3$\uparrow$ & F AvgR$\downarrow$ \\
\hline
\textbf{Ours} & \textbf{79.17} & \textbf{89.17} & 92.92 & \textbf{1.68} & \textbf{33.20} & \textbf{50.79} & \textbf{59.09} & \textbf{12.87} \\
w/o gate & 74.79 & 87.08 & 91.04 & 1.95 & 29.45 & 44.86 & 54.74 & 16.22 \\
tanh gate & 76.04 & 89.58 & 92.29 & 1.75 & 29.84 & 46.05 & 55.34 & 13.35 \\
per-frame & 74.17 & 89.17 & \textbf{93.54} & 1.79 & 29.64 & 46.44 & 57.11 & 14.12 \\
w/o $\mathcal{L}_{\mathrm{anc}}$ & 76.25 & 87.92 & 92.71 & 1.89 & 27.27 & 43.48 & 55.14 & 15.66 \\
w/o $\mathcal{L}_{\mathrm{hf}}$ & 76.88 & 88.54 & 91.46 & 1.75 & 30.24 & 44.66 & 55.14 & 13.61 \\
w/o skip & 76.25 & 89.70 & 92.29 & 1.73 & 28.06 & 43.48 & 53.16 & 13.70 \\
w/o $\hat{x}_0$ (anchor in $v$) & 76.67 & 87.92 & 92.29 & 1.88 & 31.03 & 46.64 & 54.35 & 16.25 \\
\hline
\end{tabular}
\label{tab:ablation}
\vspace{-15pt}
\end{table}


\label{sec:experiments}

\subsection{Dataset and Evaluation Metrics}
\label{sec:dataset_metrics}

\paragraph{Datasets.}
We evaluate on MotionFix~\citep{athanasiou2024motionfix} and STANCE Adjustment from OmniME~\citep{shi2026omni}. MotionFix provides 6,730 (source, instruction, target) triplets: 5,387 for training, 330 for validation, and 1,013 for testing, at 30\,fps. STANCE is a separate 20\,fps benchmark with 4,411 triplets.

\paragraph{Motion retrieval.}
We use a pretrained TMR encoder~\citep{petrovich2023tmr} to rank reference target motions by their embedding similarity to each generated motion. We report R@1, R@2, R@3, and average rank for batch retrieval and full-pool retrieval as previous methods~\citep{athanasiou2024motionfix,shi2026omni}. Higher recall and lower average rank indicate better performance.


\paragraph{Implementation details.}
We build on SimMotionEdit~\citep{li2025simmotionedit}, using a frozen CLIP ViT-L/14 text encoder and a 207-dimensional motion representation at 30 fps. We initialize SuperMotion from the trained backbone and jointly train the denoiser and preservation head using AdamW with a learning rate of $10^{-4}$ and a batch size of 128 on a single NVIDIA A100 GPU. We use velocity prediction with a squared-cosine noise schedule and 300 diffusion steps. At inference, both source and text guidance scales are set to 3.0, and source anchoring is applied after guidance at every reverse step.

\subsection{Comparison with State-of-the-Art}
\paragraph{Quantitative Comparison.} Table~\ref{tab:motionfix} and Table~\ref{tab:stance} report quantitative comparisons. On MotionFix, our method ranks first on seven of eight metrics, with OmniME~\citep{shi2026omni} marginally ahead on \textbf{FR@3}. A more informative reference is the \emph{copy source} baseline, which returns the input unchanged and still reaches $74.79\%$ \textbf{BR@1}. Most prior editors fall below it, indicating that regenerating the target from noise destroys more source content than the edit itself introduces. Only OmniME and our method exceed this baseline, and ours does so by the larger margin on both batch ($+4.38$) and full-pool ($+2.96$) R@1. On STANCE Adjustment, our method leads all batch-retrieval metrics and improves full-pool AvgR from $22.77$ to $19.21$. \textbf{FR@1} is lower than OmniME, while \textbf{FR@3} is higher and full-pool AvgR is lower, indicating fewer severe ranking failures rather than more top-1 matches. Across the two benchmarks, SuperMotion is the only method that exceeds the copy-source baseline on every retrieval metric on both benchmarks. 

\paragraph{Qualitative Comparison.} 
Figure~\ref{fig:qual_motionfix} presents qualitative comparisons on MotionFix. The first two examples assess instruction compliance. For \emph{move the hands slower}, the baselines keep the arm moving at the source speed and leave its trajectory unchanged,  whereas our result reduces the hand velocity while keeping the arm path and body configuration of the source. For \emph{spread legs less, look around and rapidly look behind you at the end}, which combines a continuous adjustment with a terminal action, only our method produces both, with the final turn matching the ground-truth timing. The third example, ``\emph{move faster}'', instead probes source preservation. The instruction changes only the temporal profile, yet TMED and SimMotionEdit resynthesize the limb trajectories and drift from the source configuration. Our result accelerates the sequence while retaining the original motion pattern, consistent with the detail-ratio behavior reported in Section~\ref{sec:introduction}. 


\subsection{Ablation Study}
\label{sec:ablation}
\paragraph{Setup.}
Table~\ref{tab:ablation} ablates each component on MotionFix. All variants are trained with the same schedule and seed. 

\paragraph{Effectiveness of the preservation gate.}
Fine-tuning the base model for the same schedule without the preservation gate $g$ (``\emph{w/o gate}'' in Table~\ref{tab:ablation}) reduces \textbf{FR@1} from 33.20 to 29.45 and raises \textbf{F AvgR} from 12.87 to 16.22. The improvement, therefore, stems from the proposed mechanism rather than from additional optimization. To verify the effectiveness of the range of the preservation gate, we replace the sigmoid gate \(g=\sigma(\cdot)\in(0,1)\) with \(\tanh\), so \(g\in(-1,1)\) (``\emph{tanh gate}'' in Table~\ref{tab:ablation}).
Positive \(g\) still copies the source; negative \(g\) pushes \(\hat{x}_0\) away from it.
For ``tanh gate'' \textbf{FR@1} falls to 29.84, near the no-gate control (``w/o gate'' with 29.45).
The gate has to be one-sided. Restricting the correction to preservation, rather than introducing an additional degree of freedom, is therefore essential.  In our framework, we predict a per-frame, per-dimension gate \(g\in(0,1)^{T\times D}\).
``per-frame'' replaces it with one scalar per frame, broadcast across all pose channels, so every joint shares the same copy strength at time \(t\).
This run reaches 29.64 \textbf{FR@1}, 3.56 below Ours (33.20),
and trails on every full-pool metric, consistent with the learned gate keeping whole feature channels high or low across time, which a per-frame scalar cannot express.

\paragraph{Effectiveness of training objectives.}
``\emph{W/o $L_{\mathrm{hf}}$}'' turns off the second-difference term in total objective.
\textbf{FR@1} falls from 33.20 to 30.24, and batch R@1 from 79.17 to 76.88. These results verify that high-frequency loss is able to help model generate accurate motion instructed by input text. Removing the anchor loss (\emph{w/o $\mathcal{L}_{\mathrm{anc}}$}) is more harmful. The gate retains the capacity to reuse the source but receives no supervision on which coordinates admit reuse, since $\mathcal{L}_v$ does not propagate to it.
\textbf{BR@1} stays at 76.25 while \textbf{FR@1} drops to 27.27, below the model without any gate value (29.45).

\begin{wrapfigure}{r}{0.5\textwidth}
    \centering
    \vspace{-10pt}
    \includegraphics[width=\linewidth]{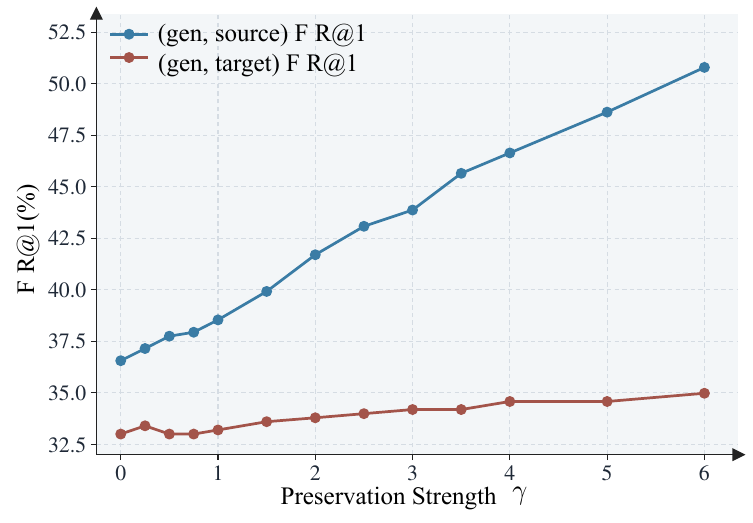}
    \vspace{-20pt}
    \caption{Preservation strength at inference. Scaling the learned gate by $\gamma$ without retraining raises retrieval against the source far more than against the target.}
    \label{fig:abla}
    \vspace{-15pt}
\end{wrapfigure}

\paragraph{Structure Evaluation.}
Removing the U-ViT skips (``\emph{w/o skip}'' in Table~\ref{tab:ablation}) while keeping the gate lowers \textbf{BR@1} from 79.17 to 76.25 and \textbf{FR@1} from 33.20 to 28.06. The skips carry high-frequency motion features directly to the late blocks, and the preservation head depends on those features to decide where reuse is acceptable.



\paragraph{Preservation strength.}
We scale the learned gate by $\gamma$ at inference without retraining (Figure~\ref{fig:abla}). At $\gamma=0$ the anchor is removed, yet \textbf{FR@1} against the target stays at 33.0, close to 33.20 at $\gamma=1$ and well above the model trained without the gate (29.45). Training through the anchor thus shapes the denoiser itself. Increasing $\gamma$ to 6 improves \textbf{FR@1} against the target by only 1.78, from 33.20 to 34.98, but improves \textbf{FR@1} against the source by 12.25, from 38.54 to 50.79. At $\gamma=6$ the generated motion retrieves its own source (50.79 \textbf{FR@1}) more accurately than its target (34.98 \textbf{FR@1}). The additional gain is thus obtained by copying rather than by editing, and we report $\gamma=1$ throughout.


\begin{figure}[ht]
    \centering
    \vspace{-10pt}
    \includegraphics[width=\linewidth]{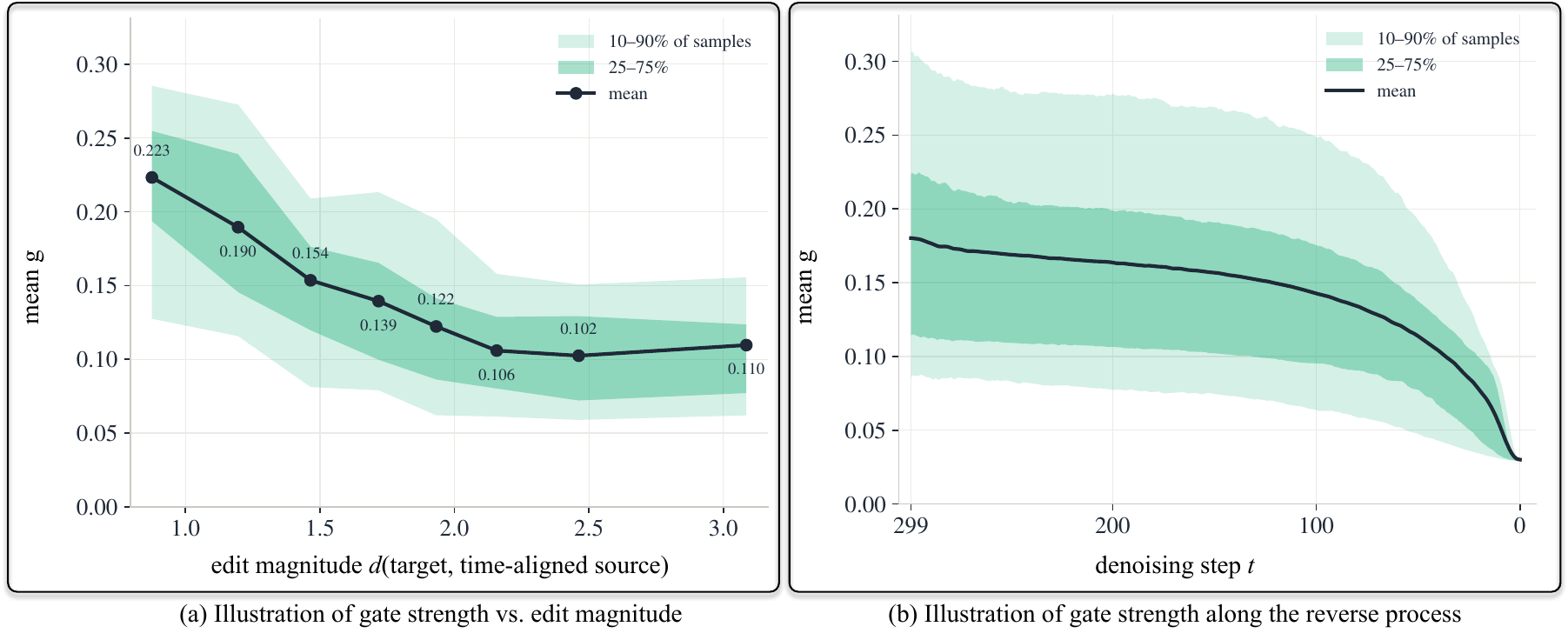}
    \vspace{-20pt}
    \caption{\textbf{Behavior of the learned gate.} (a) Mean gate value $g$ against edit magnitude $d(\text{target}, \text{time-aligned source})$, with interquartile and 10--90\% bands across samples in each bin. (b) Mean gate over frames and channels along the reverse trajectory, with the same bands.}
    \vspace{-10pt}
\label{fig:gate-learning}
\end{figure}

\paragraph{Gate Learning.}
Equation~\ref{eq:gstar} says the optimal gate should behave in two ways: it should be smaller when the edit is larger, and it should go to zero at the end of denoising. Figure~\ref{fig:gate-learning} shows the learned gate does both. In Figure~\ref{fig:gate-learning}(a), the mean gate drops from 0.223 in the smallest-edit bin to about 0.10 in the largest bins. These results verify that the gate assigns lower reuse weights to sequences that require larger modifications. In Figure~\ref{fig:gate-learning}(b), reuse stays near $0.18$ for most of the reverse process and then falls to $0.03$ over the last fifty steps. The source helps while the trajectory is still being decided, but the aligned source is not the target, so holding onto it at $t=0$ would introduce error instead of detail.
From Figure~\ref{fig:gate-learning}, without any edit mask, supervision from the editing target alone teaches the gate how much of the source to reuse and when to release the source motion.
\section{Conclusion}\label{sec:con}

In this work, we analyze the denoising process of existing methods and identify the temporal-detail attenuation as a systematic failure of diffusion motion editors. We propose Source-Preserving Denoising, which makes source reuse an explicit operation at every reverse step by blending the clean-motion estimate toward the aligned source through a learned per-frame, per-dimension gate. Extensive experiments demonstrate that SuperMotion consistently outperforms state-of-the-art editors on both MotionFix and STANCE Adjustment. Ablations attribute the improvement to the learned preservation gate, which enables selective source reuse and preserves high-frequency temporal detail that competing methods attenuate.
\section{AI use statement}
\label{app:llm}

The manuscript was drafted by the authors. We then used Claude Opus 5.5 to polish the prose and to check typos, asking it to improve sentence-level wording while leaving the technical content unchanged. After that pass, we read the revised text ourselves and corrected every phrasing that no longer matched what we had meant. All claims, experimental choices, and reported numbers are the authors'.

\bibliography{iclr2027_conference}
\bibliographystyle{iclr2027_conference}
\clearpage
\appendix

\section{Evaluation Protocol}
\label{app:pools}

\paragraph{MotionFix retrieval.}
The generation loader writes 836 sequences, consisting of a random half of the official test split (506 examples) and the 330 validation examples. The TMR gallery~\citep{athanasiou2024motionfix} is the evaluator's test set, so full-pool metrics rank each query against the \emph{506-id test pool}, not against all 836 generated files. Validation outputs never enter that gallery. Batch retrieval sorts the same IDs, shuffles them, and forms disjoint batches of 32, leaving 26 examples unused (480 scored queries). A batch percentage is therefore not a statistic over all 506 examples.

Published MotionFix numbers use this pool. Table~\ref{tab:motionfix} reports SuperMotion at 33.20 full-pool R@1.

\paragraph{STANCE Adjustment.}
The published protocol ranks against 196 IDs. OmniME is quoted from its paper. TMED, SimMotionEdit and MotionRefit have no official STANCE checkpoint, so those rows in Table~\ref{tab:stance} are our runs of their official code.

\section{Baseline Implementations}
\label{app:baselines}

MotionFix rows in Table~\ref{tab:motionfix} follow the numbers reported by each paper. The notes below record only the representation conversions used for the detail-ratio and copy-line measurements.

\paragraph{TMED and SimMotionEdit.}
Official checkpoints, evaluated with their released loaders at guidance 2.0.

\paragraph{MDM.}
A pretrained text-to-motion model conditioned on the edit instruction only. HumanML3D features are converted to joints, fitted to SMPL, and aligned into the MotionFix world frame.

\paragraph{MDM-BP.}
The same MDM backbone, with GPT body-part labels. Unmentioned parts are copied from the source in feature space at every reverse step, so the median detail ratio of $1.01\times$ is obtained by construction.

\paragraph{MotionRefit.}
Evaluated zero-shot from their STANCE-trained weights. We convert the source to their SMPL-X 20\,fps format, fix the task to regeneration with the raw instruction, and align frames and the floor both ways.

\paragraph{MotionLab.}
Treated as one task of their unified generative model. Table~\ref{tab:motionfix} quotes their SourceTextMetrics, which do not report a full-pool protocol.

\paragraph{OmniME.}
Quoted from the paper on both benchmarks. The detail-ratio and copy-line entries in Table~\ref{tab:copy} use their official MotionFix checkpoint.

\section{Additional Metrics}
\label{app:metrics}

\paragraph{Representation.}
The denoiser operates on the 207-dimensional MotionFix feature of Section~3.1. The preservation gate and the clean-space source anchor are applied in this space. Exported TMR features drop the 22 local joint coordinates and convert increments to absolute SMPL parameters, yielding 135 dimensions (translation 3, global orientation 6D, body pose 126). Distances in Table~\ref{tab:copy} use the 135-d export.

\paragraph{Detail ratio.}
Section~1 defines the detail ratio on the running clean-motion estimate during denoising. Dataset-level numbers in Table~\ref{tab:copy} use the same second-order difference magnitude on the exported 135-d pose, aggregated as the median over $N{=}836$ generated files. A value of $1\times$ matches the target's second-difference magnitude but does not imply a correct motion. MotionLab's median is $1.40\times$. The $1.01\times$ of MDM-BP is produced by copying source features.

\paragraph{Copy source.}
Let $d(\cdot,\cdot)$ be the mean per-frame $L_2$ distance on the 135-d pose. Time-aligning the source onto the target timeline and scoring that copy gives $d(s,x_0)=1.745$. Any editor whose $d(\hat{x}_0, x_0)$ exceeds $1.745$ ends farther from the target than the copy-source baseline, as TMED does at $1.885$. SuperMotion sits at $1.586$, with $54.8\%$ of outputs closer to the source than to the target, which is less source-biased than OmniME ($60.8\%$) and TMED ($60.9\%$). This is the same control as Figure~\ref{fig:abla}. Raising the gate scale improves retrieval by copying, not by editing.

\begin{table}[t]
\centering
\small
\caption{Copy-versus-edit diagnostic on MotionFix ($N{=}836$).}
\label{tab:copy}
\begin{tabular}{lcccc}
\toprule
Method & Detail ratio & $d(\mathrm{gen},\mathrm{tgt})$ & $d(\mathrm{gen},\mathrm{src})$ & Closer to src \\
\midrule
TMED & 0.330 & 1.885 & 1.791 & 60.9\% \\
SimMotionEdit & 0.708 & 1.661 & 1.616 & 56.1\% \\
OmniME & 0.658 & 1.642 & 1.496 & 60.8\% \\
SuperMotion & 1.035 & 1.586 & 1.563 & 54.8\% \\
\midrule
Copy source & -- & 1.745 & 0 & -- \\
\bottomrule
\end{tabular}
\end{table}

\section{Sampling Procedure}
\label{app:algorithm}

\begin{center}
\fbox{\begin{minipage}{0.94\linewidth}
\textbf{Algorithm 1.} Source-preserving reverse step.\\[3pt]
\textbf{Input.} Source $s$, instruction $c$, length $T$, trained denoiser and preservation head, noise schedule.\\
1. Align $\bar s\leftarrow\mathrm{Align}(s,T)$ by linear resampling. Draw $x_T\sim\mathcal N(0,I)$.\\
2. For $t=T,\ldots,1$, repeat the following steps.\\
\hspace*{1em}(a) Evaluate the unconditional, source-only, and source-plus-text branches at the same $x_t$.\\
\hspace*{1em}(b) Form the guided velocity $\hat v_{\mathrm{cfg}}$ by Eq.~(3).\\
\hspace*{1em}(c) Convert $\hat x_0$ by Eq.~(2).\\
\hspace*{1em}(d) Read $g=\sigma(f_\phi(h))$ from the fully conditioned branch.\\
\hspace*{1em}(e) Set $\tilde x_0=\hat x_0+g\odot(\bar s-\hat x_0)$ by Eq.~(6).\\
\hspace*{1em}(f) Sample the DDPM posterior $q(x_{t-1}\mid x_t,\tilde x_0)$. Do not convert the correction back to velocity.\\
3. Return the final clean motion.
\end{minipage}}
\end{center}

Guidance is applied to the velocity branches before the blend. The gate is not itself guided, because extrapolating $g$ would break the convex-combination interpretation of Eq.~(6). All three branches are evaluated at the same $x_t$, since a replica mismatch under velocity prediction produces an inconsistent reverse step.

\section{Preservation-Head Architecture}
\label{app:impl}

Section~4.1 gives the training setup. The preservation head is a three-layer MLP on the same target-token features $h\in\mathbb{R}^{T\times512}$, with widths $512\rightarrow1024\rightarrow1024\rightarrow207$ and SiLU activations. The last layer is zero-initialized with bias $-6$, so $g=\sigma(-6)\approx0.00247$ and the sampler matches the parent model at initialization. The velocity loss is applied to the ungated prediction, while $\mathcal L_{\mathrm{anc}}$ and $\mathcal L_{\mathrm{hf}}$ are applied to $\tilde x_0$.





\begin{figure}[ht]
\centering
\includegraphics[width=\linewidth]{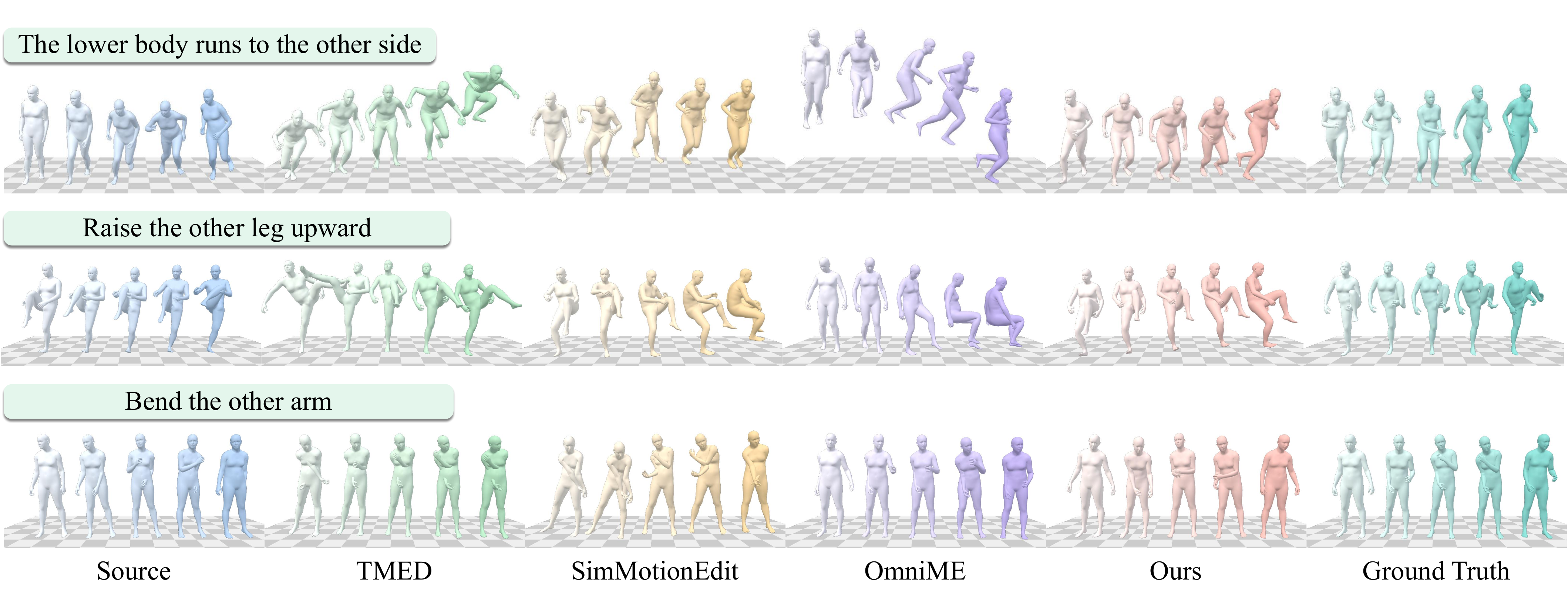}
\caption{Qualitative comparison on STANCE Adjustment. Each instruction targets a specific limb or side. Left to right: source, TMED, SimMotionEdit, OmniME, SuperMotion, and the reference target.}
\label{fig:qual_stance}
\end{figure}

\section{Qualitative Results on STANCE Adjustment}
\label{app:qual_stance}

Figure~\ref{fig:qual_stance} shows results on STANCE Adjustment, where each instruction targets a specific limb or side. In \emph{the lower body runs to the other side}, the baselines introduce a full-body reorientation or an airborne pose absent from the source, while our result redirects the lower body alone. In \emph{raise the other leg upward} and \emph{bend the other arm}, the baselines frequently act on the wrong limb or apply the change to both sides. Our method selects the intended one and leaves the remaining joints close to the source.

\end{document}